\documentclass[lettersize,journal]{IEEEtran}
\usepackage{amsmath,amsfonts}
\usepackage{algorithmic}
\usepackage{algorithm}
\usepackage{array}
\usepackage[caption=false,font=normalsize,labelfont=sf,textfont=sf]{subfig}
\usepackage{textcomp}
\usepackage{stfloats}
\usepackage{url}
\usepackage{verbatim}
\usepackage{graphicx}
\usepackage{cite}

\usepackage{times}
\usepackage{epsfig}
\usepackage{amssymb}
\usepackage{enumitem}
\usepackage{booktabs}
\usepackage{pifont}
\usepackage[super]{nth}

\newcolumntype{C}[1]{>{\centering\arraybackslash}p{#1}} 
\newcommand{\xmark}{\ding{55}}%

\begin{document}

\title{LEST: Large-scale LiDAR Semantic Segmentation with Transformer}

\author{Chuanyu Luo, Nuo Cheng, Sikun Ma, Han Li, Xiaohan Li, Shengguang Lei, Pu Li
\thanks{Chuanyu Luo, Nuo Cheng are with the LiangDao GmbH, Berlin, 12099, Germany and also with the Ilmenau University of Technology, Ilmenau, 98693, Germany (email: chuanyu.luo@tu-ilmenau.de; nuo.cheng@tu-ilmenau.de).}
\thanks{Sikun Ma, Han Li, Xiaohan Li, Shengguang Lei are with the LiangDao GmbH, Berlin, 12099, Germany (email: sikun.ma@liangdao.de; han.li@liangdao.de;xiaohan.li@liangdao.de;shengguang.lei@liangdao.de)}
\thanks{Pu Li is with the Ilmenau University of Technology, Ilmenau, 98693, Germany (email: pu.li@tu-ilmenau.de).}}

\markboth{Journal of \LaTeX\ Class Files,~Vol.~14, No.~8, August~2021}%
{Shell \MakeLowercase{\textit{et al.}}: A Sample Article Using IEEEtran.cls for IEEE Journals}

\IEEEpubid{0000--0000/00\$00.00~\copyright~2021 IEEE}

\maketitle

\begin{abstract}
Large-scale LiDAR-based point cloud semantic segmentation is a critical task in autonomous driving perception. Almost all of the previous state-of-the-art LiDAR semantic segmentation methods are variants of sparse 3D convolution.  Although the Transformer architecture is becoming popular in the field of natural language processing and 2D computer vision, its application to large-scale point cloud semantic segmentation is still limited. In this paper, we propose a \textbf{L}iDAR s\textbf{E}mantic \textbf{S}egmentation architecture with pure \textbf{T}ransformer, LEST. LEST comprises two novel components: a Space Filling Curve (SFC) Grouping strategy and a \textbf{Dis}tance-based \textbf{Co}sine Linear Transformer, DISCO. On the public nuScenes semantic segmentation validation set and SemanticKITTI test set, our model outperforms all the other state-of-the-art methods.
\end{abstract}


\section{Introduction}
\IEEEPARstart{I}{n} an autonomous driving system, LiDAR-based point cloud 3D environment perception is important for safe and reliable driving. Unlike image-based 2D perception tasks, the large-scale point cloud is irregular, sparse and unordered. The 3D environment perception includes tasks such as 3D object detection and point cloud semantic segmentation.

Unlike the 3D object detection task, the 3D semantic segmentation task usually requires more granular and spatial information, and these requirements make the semantic segmentation task more challenging.

In deep learning-based 3D perception approaches, the pioneering work PointNet~\cite{qi2017pointnet} is the first to aggregate the local unordered points features by a symmetric function, max-pooling. PointPillars~\cite{lang2019pointpillars} applies a simple PointNet to each pillar and uses it to learn a representation of point clouds in a pillar. The pillars are then mapped into a 2D Bird’s-Eye-View (BEV). A series of dense 2D convolution layers is further used for 3D object detection. However, mapping 3D objects to 2D BEV could result in significant information loss, especially for small objects in the semantic segmentation task.

\begin{figure}[h]
\begin{center}
\includegraphics[width=0.8\linewidth]{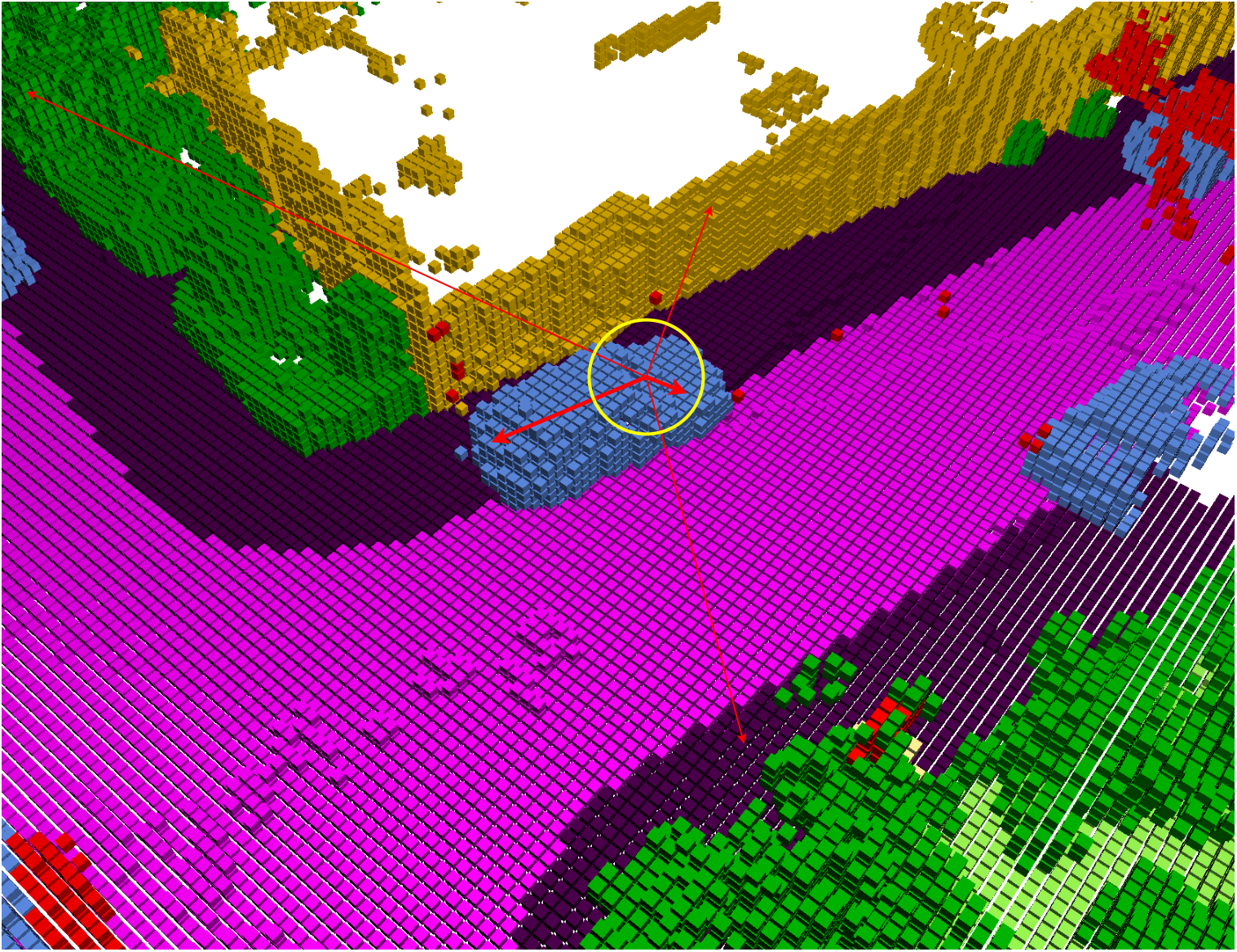}
\end{center}
   \caption{The comparison of self-attention and convolution for voxels. The point cloud is at first voxelized. The red directional arrows indicate the attention between a query voxel on a car and all the other voxels. A thicker arrows represent a higher attention. In this example, the query voxel pays more attention to the voxels on the same car, and less attention to the background voxels. The yellow circle represents a convolutional limited receptive field, which is not enough for huge objects like the trucks. Unlike the convolution, the self-attention mechanism has global receptive field.}
\label{fig:voxel_attenion}
\end{figure}

Traditional dense 3D convolution is inefficient for processing 3D sparse data. In 3D object detection, SECOND~\cite{yan2018second} introduces a sparse 3D convolution operator to address this issue. Following this, Polar-Coordinate-System-based Cylinder3D~\cite{Zhu_2021_cylinder3d} and Neural-Architecture-Search-based SPVNAS~\cite{spvnas} apply the sparse 3D convolution to the 3D semantic segmentation task and achieve state-of-the-art results.

\IEEEpubidadjcol

Although Transformer~\cite{vanilla_transformer} is dominant in natural language processing (NLP), and has become popular in the image-based 2D computer vision field, its application to large-scale 3D point cloud is still limited. However, a voxel in 3D perception has a similar representation as a word in NLP, as both can be generalized as a token with high-dimensional features learned through training. The comparison of self-attention and convolution for voxels is shown in Figure~\ref{fig:voxel_attenion}.

One of the challenges when applying the self-attention mechanism to voxels/tokens is the extremely large number of voxels. The vanilla Transformer has quadratic complexity in terms of the number of voxels. To address this issue, inspired by the Swin-Transformer~\cite{liu2021swinTransformer} in image tasks, SST~\cite{fan2022_SST} separates the voxels into rectangle windows, and the complexity is reduced by using self-attention only within each window. The shifted window method is then used to expand the receptive field across different windows.

However, one limitation of the SST is that the number of voxels in each fixed-size window is significantly different due to the varying density of the point cloud. This variation in the number of voxels can lead to inefficiencies in parallel training and inference, as well as increased memory usage. Furthermore, the shifted window-based method can be generalized as an extension of ensemble models, and the key advantage, global receptive field, from Transformer is not theoretically guaranteed. 

In long range sequences tasks in NLP, where thousands of words are processed simultaneously, linear Transformer methods are more popular. Linear Transformers have only linear complexity in the number of tokens and have a theoretical global receptive field. The key idea of Linear Transformers is to decompose the softmax operator of the self-attention module into a linear form~\cite{zhen2022cosformer}.

In our paper, we propose a Space Filling Curve (SFC) Grouping strategy to efficiently separate the voxels into multiple groups and aggregate the local voxels features in each group by a downstream vanilla Transformer. Additionally, we propose a novel linear Transformer to build a global receptive field with only linear complexity and strong representation ability.

Our contributions can be summarized as follows:

1. We propose a voxel-based Transformer-based 3D backbone for LiDAR semantic segmentation task, and achieves impressive results compared with the other state-of-the-art method.

2. A novel SFC Grouping strategy is proposed, and the voxel local features can be aggregated within a group. It is proved that, the combination of our grouping strategy with the  vanilla Transformer has the lowest expected value of the complexity.

3. We propose a novel Linear Transformer method with global receptive field but only linear complexity. Linear Transformer is popular in NLP especially for long range sequences task. As far as we know, we are the first to unify the 3D perception task in Computer Vision with the long range sequences task~\cite{tay2021lra} in NLP. The proposed unified method can reduce the domain gaps between CV and NLP research.

\section{Related work}
\label{sec:r_work}
\subsection{Large-scale point cloud semantic segmentation.} In the large-scale point cloud semantic segmentation task, the mainstream approaches include point-based methods, projection-based methods and voxel-based methods.

\subsubsection{Point-based method.} Most point-based methods pipeline includes point sampling, neighbors searching, features aggregation and classification~\cite{randLA, Luo2022A03_mvpNEt}. One key disadvantage of point-based methods is that the inefficient neighbor searching method like K-Nearest Neighbors (KNN) is recursively used. Although MVP-Net~\cite{Luo2022A03_mvpNEt} replaces the KNN method by Space Filling Curves for high efficiency, the performance of point-based methods is still limited compared to voxel-based methods. 

\subsubsection{Projection-based method.} To leverage the success of 2D images, projection-based methods map the 3D points to a 2D pseudo-image, aggregate the features from neighboring pixels, and then inversely map the pixels to the 3D point cloud. The projection-based methods~\cite{wu2019squeezesegv2, milioto2019rangenet++} map the point cloud to a spherical projection, and the PolarNet~\cite{zhang2020polarnet} maps the point cloud to a polar BEV. However, the information loss due to the 3D-to-2D projection limits the performance of projection-based methods.

\subsubsection{Voxel-based method.} Voxel-based methods~\cite{Zhu_2021_cylinder3d}~\cite{spvnas}~\cite{yan2021JS3C-Net} use PointNet to learn voxel representations from points within each voxel. The voxels features are then aggregated using Sparse 3D Convolution~\cite{yan2018second, 3DSemanticSegmentationWithSubmanifoldSparseConvNet}, which is efficient for sparse data and incorporates priori knowledge of the voxel and its neighbors. In large-scale scene, 3D convolution has limited receptive field and cubic complexity on the size of the convolution kernel.

\subsection{Space filling curves grouping.}\label{sec:spc_r_work} 
\begin{figure}[h]
\begin{center}
\includegraphics[width=0.6\linewidth]{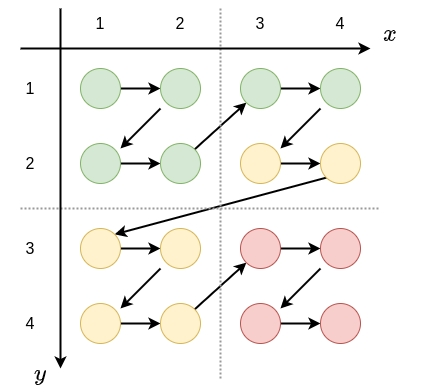}
\end{center}
   \caption{Morton-order in dense 2D data points. Each color represents one group. The 2D spatial locality from points is preserved after mapping to a 1D sequence. Along the sorted points, every 6 point are separated into a group. Except for the last group, all the other groups have the same number of data points. In our 3D perception task, the data points represent voxels, and we apply the same sorting and grouping strategy as used for dense data points to the sparse 3D voxels.}
\label{fig:morton_code}
\end{figure}

Space filling curves (SFC) is a sorting method to map the high dimensional data to one dimensional sequence while preserving the locality of the data points~\cite{Thabet_2020_CVPR_Workshops_morton}. One of the widely used and high efficient SFC method is the Morton-order~\cite{morton1966computer}, also known as Z-order because of the curve shape in the 2D case. Along the sorted by SFC sequences, the data points can be separated into different groups efficiently, and all the groups have almost the same number of data points, as illustrated in Figure~\ref{fig:morton_code}.

\subsection{Transformers in vision tasks.} Transformer~\cite{vanilla_transformer} is firstly proposed in the NLP field. In the 2D computer vision tasks, ViT~\cite{dosovitskiy2021an_vit} splits the image into patches and then uses the vanilla Transformer. PVT~\cite{wang2021pvt} is the first hierarchical design for ViT and is used in various dense prediction tasks like 2D object detection and semantic segmentation tasks. Swin Transformer~\cite{liu2021swinTransformer} is a multi-stage hierarchical architecture, and use Transformer in gradually shifted windows to extend the pixels receptive field. 

\begin{figure*}[h]
\begin{center}
\includegraphics[width=0.8\linewidth]{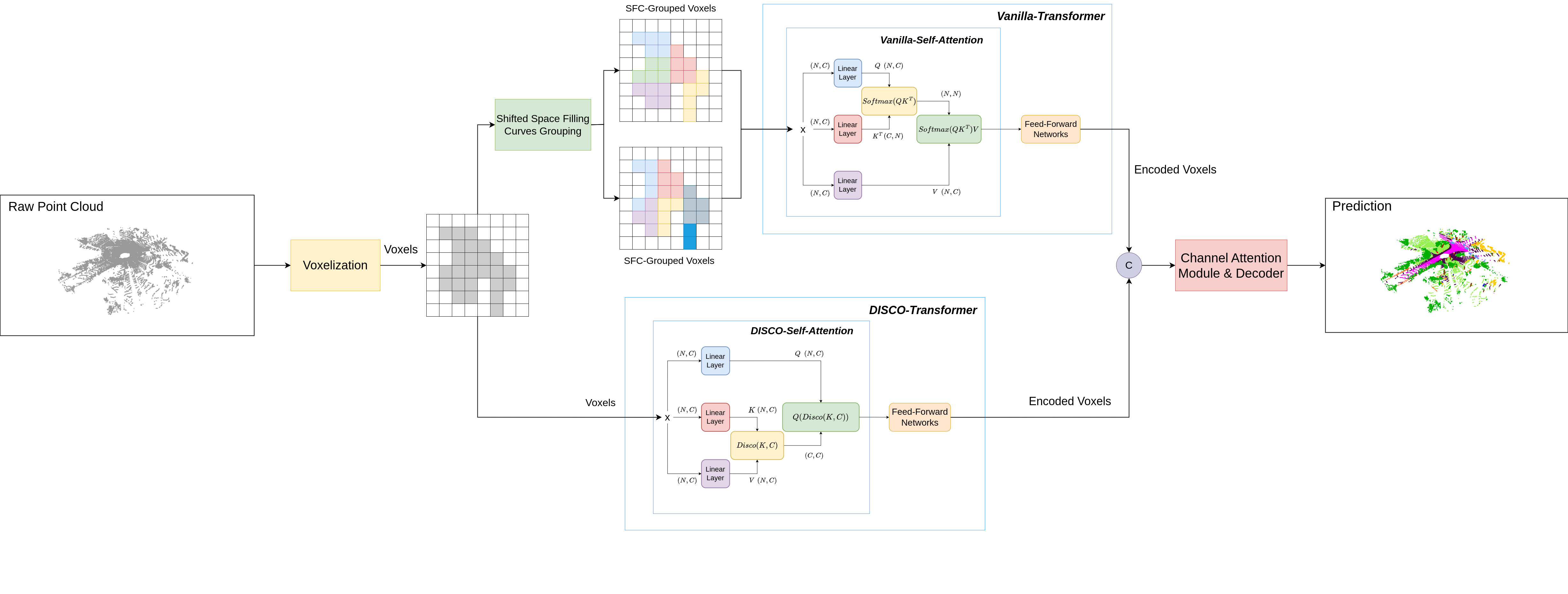}
\end{center}
\caption{The architecture of our proposed model LEST. The raw point cloud is at first voxelized with one mini-PointNet~\cite{qi2017pointnet}. For illustration purpose, the sparse 3D voxels are represented as sparse 2D pixels here. In the up branch, the voxels are grouped by Space Filling Curves (SFC), as introduced in section~\ref{sec:spc_r_work}. In each grid, a color indicates a group. Due to different grouping strategies, the same voxel can be grouped into different groups in different grids. This mechanism is an extension of the shifted window methods such as~\cite{liu2021swinTransformer, fan2022_SST}. SFC-Grouping ensures that near voxels are grouped together, and the number of voxels in each group is almost the same. The downstream vanilla Transformer can be efficient  because of the balanced voxel distribution. In the down branch, a novel Linear Transformer is proposed, Distance Cosine Linear Transformer, to build a theoretical global field of voxels with only linear complexity. Then, the output feature-encoded/aggregated voxels from the two branches are concatenated. After being decoded by a simple channel attention module~\cite{chaneel_attention} and an MLP-based decoder, the semantic label per point is predicted.}
\label{fig:architecture}
\end{figure*}

In the 3D object detection task, VoTr~\cite{mao2021voxel_votr} proposes the first Transformer-based model. In VoTr, a GPU-based hash table is used to search neighboring voxels, with each voxel serving as a query in the self-attention module. Most related work to our proposed SFC-Grouping method is the  SST~\cite{fan2022_SST}, a 3D object detection architecture. SST firstly pillarize the LiDAR points, and then gradually shifted window-based groups the pillars. Transformer is used in each group to aggregate the pillar features. However, window-based method requires extra high memory usage and not feasible in 3D semantic segmentation task. Another problem is that though the group is gradually shifted, the tokens can still not have a real global receptive field.

\subsection{Transformer with linear complexity}

Dot-product attention with softmax normalization in Transformer self-attention module is the key to have long range dependency and global receptive field. However, the quadratic complexity of self-attention module makes it impossible to long sequence tasks in NLP, or 3D semantic segmentation tasks that include thousands of voxels.

Recently, many works are proposed to make the Transformer more efficient and has only linear complexity. Kernel-based linear Transformer~\cite{katharopoulos2020linear_transformer} uses kernel function to approximate softmax normalization to linearize the computation in self-attention. SOFT~\cite{lu2021soft} propose a softmax-free Transformer and use the Gaussian kernel function to replace the dot-product similarity. The most related work to our proposed DISCO is CosFormer~\cite{zhen2022cosformer}, which replaces the softmax operator by two attention properties: non-negativeness and non-linear re-weighting scheme. Like the vanilla Transformer, CosFormer still uses the dot-product as tokens similarity. In our proposed DISCO module, the similarity is the 1-norm distance between tokens and is showed to have better performance than dot-product similarity in CosFormer in ablation studies.

.

\section{Methodology}
To tackle the semantic segmentation tasks in a large-scale LiDAR-based scenario, we propose a pure Transformer-based architecture called LEST. LEST includes two novel components: a \textbf{S}pace \textbf{F}illing \textbf{C}urves (SFC) Grouping Transformer, which is proposed to build voxels internal interaction within a group, and a \textbf{Dis}tance \textbf{Co}sine linear Transformer (DISCO), which is proposed to have one global receptive field across groups. The whole architecture is shown in Figure~\ref{fig:architecture}.

\subsection{Space Filling Curves Grouping}
\begin{figure}[h]
\begin{center}
\includegraphics[width=0.75\linewidth]{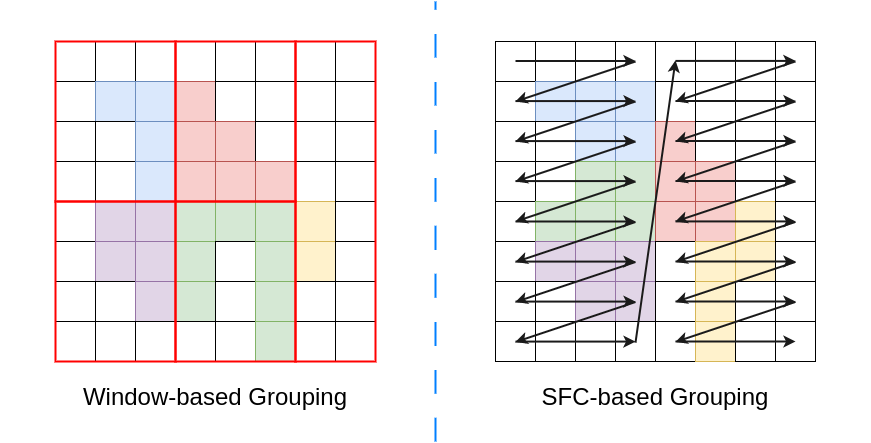}
\end{center}
   \caption{A comparison of the window-based grouping method and SFC-based grouping method. The red rectangles in the left grid represent the windows in the window-based grouping method. Note that in almost all window-based methods~\cite{fan2022_SST, liu2021swinTransformer}, the size and number of windows is fixed. In the right SFC-based grouping method, the black arrows is the space filling curves introduced in section~\ref{sec:spc_r_work}. The voxel distribution in groups in SFC-based grouping method is more balanced.}
\label{fig:window_vs_sfc}
\end{figure}

In section~\ref{sec:spc_r_work}, we introduced the Space Filling Curves (SFC) method, which is used to sort high-dimensional data as a 1D sequence. In this work, we use the SFC method to group nearby voxels together. Figure~\ref{fig:window_vs_sfc} shows a comparison between the commonly used Window Grouping method~\cite{fan2022_SST} and our proposed SFC Grouping method. After the voxels are grouped, a vanilla Transformer is used for each group. By using the shifted grouping method, the receptive field of a voxel is expanded.

\begin{figure}[h]
\begin{center}
\includegraphics[width=1.0\linewidth]{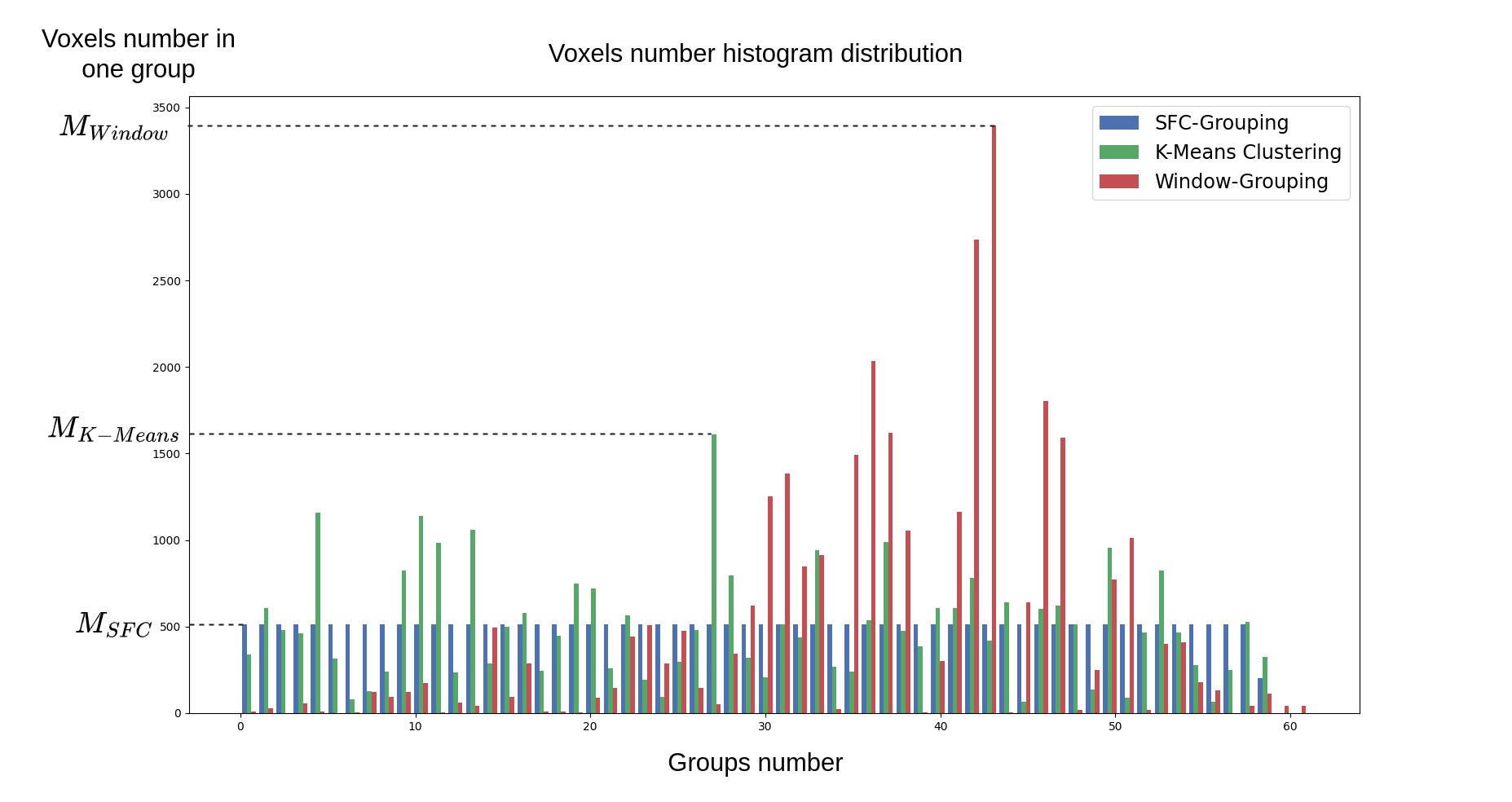}
\end{center}
   \caption{An example of LiDAR voxel distribution. Compared to K-Means-clustering grouping and Window grouping, the SFC grouping distribution is more balanced. As a result, the complexity of the downstream Transformer is significantly reduced.}
\label{fig:voxel_distribution}
\end{figure}

An advantage of the SFC grouping method is that it ensures each group contains a similar number of voxels. In contrast, previous grouping methods such as window-based grouping or K-Means clustering grouping can result in unbalanced voxel distributions among groups. One example is shown in Figure~\ref{fig:voxel_distribution}.

\subsubsection{Complexity analysis if sequentially processing}

The complexity if processing with the vanilla Transformer is analyzed here. Let $N$ denotes the number of all voxels, and $G$ indicates the number of groups. For any grouping method, $X$ is an random variable indicating the number of voxels in a group. If the number of groups $G$ is enough large, obviously we have this equation $E(X) \approx ~\frac{X_1 + X_2 + \cdots + X_G}{G} = \frac{N}{G}$.

The vanilla Transformer has quadratic complexity on the number of tokens in one group as $O(X^2)$. If we make the vanilla Transformer sequentially process all groups, the complexity is $O(X_{1}^{2} + X_{2}^{2} + \cdots + X_G^2)$. With Transformer, the expected value of complexity if using any grouping method is $O_{any} = E(X_1^{2} + X_2^{2} + \cdots + X_G^{2}) = G \cdot E(X^{2})$. As SFC grouping method guarantees that all groups has almost the same number of voxels as $\frac{N}{G}$. The expected value of complexity if using SFC grouping is $O_{SFC} = G \cdot (\frac{N}{G})^2 = G \cdot E^2(X)$. For any random variable $X$, it is not hard to prove that $E(X^2) = Var(X) + E^2(X)$. Here $Var$ is the variance. The complexity difference is shown in Equation~\ref{eq:complexity_difference}

\begin{equation}
\label{eq:complexity_difference}
\begin{aligned}
O_{any} - O_{SFC} &= G \cdot E(X^{2}) - G \cdot E^2(X) \\
                   &= G \cdot Var(X) + G \cdot E^2(X) - G \cdot E^2(X) \\
                   &= G \cdot Var(X) \geqslant 0.
\end{aligned}
\end{equation}

From Equation~\ref{eq:complexity_difference}, it can be observed that any grouping method has equal or higher complexity than the SFC grouping method. The more unbalanced the grouping strategy is, the higher the expected complexity of the downstream vanilla Transformer module.

\subsubsection{Complexity analysis if parallelly processing}

Instead of sequentially processing data with the Transformer, parallelly processing can be more efficient on GPU. In this parallel case, the voxels in each group are padded to match the maximum number of voxels in all groups.

Let $M$ denote the maximum number of voxels in a group. The complexity of the vanilla Transformer now is $GM^2$. Using SFC grouping, $M_{SFC} \approx \frac{N}{G}$, and the complexity is approximately only $O_{SFC} = GM_{SFC}^2  \approx G\frac{N^2}{G^2} = \frac{N^2}{G}$. Compared to the method without grouping, the complexity is reduced evidently from $N^2$ to $\frac{N^2}{G}$. From Figure~\ref{fig:voxel_distribution} it can be obeserved, compared to the other grouping method complexity $GM^2$ like the window-based method~\cite{fan2022_SST}, the maximum number of voxels in group $M_{Window} \gg M_{SFC}$, and the downstream Transformer complexity is also much larger than the SFC grouping method as $GM_{Window} ^2 \gg GM_{SFC}^2  \approx \frac{N^2}{G}$. 

Note that in window-based method SST~\cite{fan2022_SST} in 3D object detection task, it is shown that much GPU memory is used. In the LiDAR semantic segmentation task, which requires more granular information with smaller voxel size and a larger number of voxels, the window-based grouping method is not feasible.

\subsection{Linear Transformer background}

The Scaled Dot Product Attention is one of the key properties of the Transformer~\cite{vanilla_transformer} model. It computes the dot product of the queries with all the keys and applies a softmax function to normalize the attention weights for each query-key pair. 

Let $x \in \mathbb{R} ^ {N \times C}$ denotes a sequence of $N$ tokens with features dimension $C$. The input sequence $x$ can be projected by there learn-able matrices $W_Q \in \mathbb{R} ^ {C \times D}$, $W_K \in \mathbb{R} ^ {C \times D}$ and $W_V \in \mathbb{R} ^ {C \times D}$ to the corresponding matrices $Q$, $K$ and $V$ as follows:

\begin{equation}
\label{eq:1}
\begin{aligned}
Q &= xW_Q \\
K &= xW_K \\
V &= xW_V.
\end{aligned}
\end{equation}

The output matrix $O \in \mathbb{R} ^ {N \times D}$ of attention module can be computed as:
\begin{equation}
\label{eq:2}
O = softmax(\frac{QK^T}{\sqrt{D}})V.
\end{equation}

\begin{figure}[h]
\begin{center}
\includegraphics[width=1.0\linewidth]{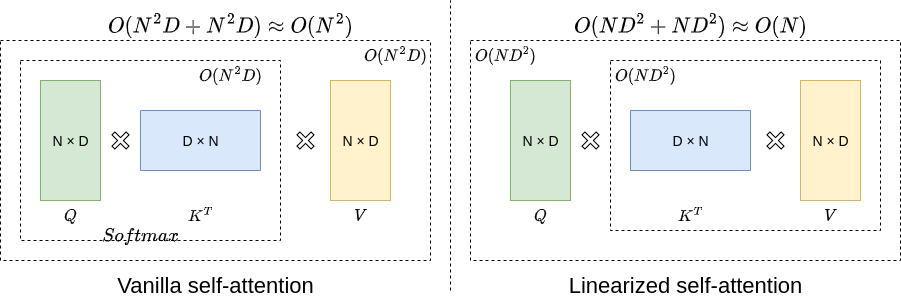}
\end{center}
\caption{Vanilla self-attention with softmax operator has complexity as $O(N^2D + N^2D)$. By removing softmax operator, and using matrix multiplication associative property, the new self-attention mechanism has complexity as $O(ND^2 + ND^2)$. In our 3D perception task, $N$ is the number of voxels and $D$ is the feature size of a voxel. As $N \gg D$, the complexity of linearized self-attention can be approximated as $O(N)$.}
\label{fig:linear_tf}
\end{figure}

It is not hard to prove that the Scaled Dot Product Attention has space and time complexity as $O(N^2D)$, which prohibits the scale-up ability if existing many tokens. By naively removing the softmax operator in Equation \ref{eq:2}, and rewriting it as Equation ~\ref{eq:2_1}

\begin{equation}
\label{eq:2_1}
\begin{aligned}
O = \frac{QK^T}{\sqrt{D}} V  = \frac{QK^TV}{\sqrt{D}} = \frac{Q(K^TV)}{\sqrt{D}},
\end{aligned}
\end{equation}
the new form $O = \frac{Q(K^TV)}{\sqrt{D}}$ has only space and time complexity as $O(ND^2)$. If $N \gg D$, the complexity of Equation~\ref{eq:2_1} is $O(N)$. The complexity of the vanilla self-attention and the linearized self-attention is further illustrated in Figure~\ref{fig:linear_tf}~\cite{zhen2022cosformer}.

However, removing the softmax function directly will cause the attention matrix elements not always be positive and not normalized.

We use $M_i$ here to represent the $i$th-row of a general matrix $M$. Equation \ref{eq:2} can be generally rewritten as:

\begin{equation}
\label{eq:general_attention_eq}
\begin{aligned}
O_i = \sum_{j=1}^{N} {\frac{Sim(Q_i, K_j)}{\sum_{j=1}^{N} Sim(Q_i, K_j)} V_j}.
\end{aligned}
\end{equation}

Here $Sim(Q_i, K_j)$ indicates the similarity of query $Q_i$ and key $K_j$. In Equation \ref{eq:2}, the similarity function of query $Q_i$ and key $K_j$ is its exponential dot product.

Previous work like Linear Transformer~\cite{katharopoulos2020linear_transformer} uses kernel function $\phi(x) = elu(x) + 1$ to approximate the softmax operator, where $elu(x)$ denotes the exponential linear unit~\cite{elu} activation function. The complete attention function is

\begin{equation}
\label{eq:general_linear_attention}
\begin{aligned}
O^{'}_i &= \sum_{j=1}^{N} {\frac{\phi(Q_i)\phi(K_j^T)V_j}{\sum_{j=1}^{N} \phi(Q_i)\phi(K_j^T)}} \\
& = \frac{\phi(Q_i)\sum_{j=1}^{N} {\phi(K_j^T)V_j}}{\phi(Q_i)\sum_{j=1}^{N} \phi(K_j^T)}.
\end{aligned}
\end{equation}
Like the Equation~\ref{eq:2_1}, the Equation \ref{eq:general_linear_attention} complexity is $O(N)$

Instead of approximating the softmax function, CosFormer~\cite{zhen2022cosformer} is based on two properties of attention matrix: non-negativeness and non-linear re-weighting ability. It proposes a decomposed similarity function with linear complexity as

\begin{equation}
\label{eq:6}
\begin{aligned}
Sim(Q_i, K_j) &= Q^{'}_i K^{'T}_jcos(\frac{\pi}{2} \times \frac{i-j}{M}) \\
Q_i^{'} &= ReLU(Q_i) \\
K_j^{'} &= ReLU(K_j). \\
\end{aligned}
\end{equation}
Here if $N$ denotes the number of all tokens, $M$ is a hyper-parameter satisfying $M \geq N$. The $cos(\frac{\pi}{2} \times \frac{i-j}{M})$ item indicates the index space distance between $Q_i$ and $K_j$. 

\subsection{Distance cosine linear Transformer}
\begin{figure}[h]
\begin{center}
\includegraphics[width=1.0\linewidth]{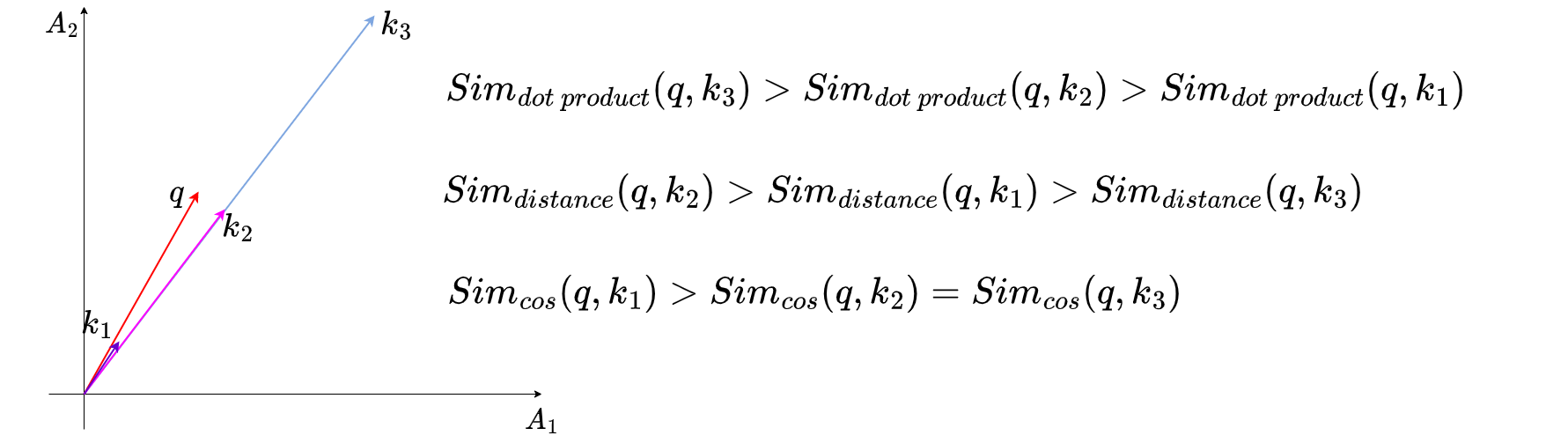}
\end{center}
   \caption{The dot product, cosine and distance similarity between vector $q$, $k_1$, $k_2$ and $k_3$. The disadvantage of cosine similarity is that it does not take the magnitude of vectors into account but only considers their direction. The dot product similarity, however, can be much sensitive to the magnitude of vector. Although the example shown is in a 2D space, the same conclusion can be extended to high-dimensional vector space.}
\label{fig:similarity}
\end{figure}

In the architecture as shown in Figure~\ref{fig:architecture}, we use a novel linear Transfomer, \textbf{Dis}tance \textbf{Co}sine Linear Transformer (DISCO), to build a voxel global receptive field. In the vanilla Transformer, the similarity between $Q_i$ and $K_j$ is the corresponding dot product like $Sim(Q_i, K_j) = Q_i \cdot K_j$. The softmax operator is then used to reweigh and normalize the similarity as attention. In addition to the dot product similarity, the cosine similarity of vectors is also widely used~\cite{zhou2021deepvit}. Instead of the dot product similarity and cosine similarity, we use the distance of vectors here as a similarity measure. Figure~\ref{fig:similarity} shows an example that distance is a better similarity measure than dot product and cosine similarity in terms of vector magnitude influence. Cosine similarity does not consider the magnitude of vectors, and the dot product similarity is not robust when one vector's magnitude is extremely large.

The commonly used distance measure is the 1-norm, the taxicab norm, or the 2-norm, the Euclidean norm. Let $D_{ij}$ denote the 1-norm distance between vector $Q_i$ and $K_j$. If $Q_i = (Q_{i1}, Q_{i2}, \cdots, {Q_{in}})$, and $K_j = (K_{j1}, K_{j2}, \cdots, {K_{jn}})$, the 1-norm distance $D_{ij}$ is

\begin{equation}
\label{eq:distance_def}
\begin{aligned}
D_{ij} = \|Q_i - K_j\|_1 = \sum_{p=1}^{n}|Q_{ip} - K_{jp}|.
\end{aligned}
\end{equation}

The similarity of vector $Q_i$ and $K_j$ can be defined as the function $Sim(Q_i, K_j) = f(D_{ij})$. The function $f$, which maps the distance to the similarity, must have at least the following two requirements:
\begin{enumerate}\label{properties}
\item The function $f$ must be a monotonically non-increasing function.
\item The output of function $f(D_{ij})$ must be positive, as a negative similarity is meaningless. The negative attention also destabilizes the Transformer performance~\cite{katharopoulos2020linear_transformer, zhen2022cosformer}.
\end{enumerate}

To make the Transformer can be decomposed and have linear complexity, the third important requirement is that the function $f$ must can be decomposed, which means

\begin{equation}
\label{eq:general_sim_and_distance}
\begin{aligned}
Sim(Q_i, K_j) = f(D(Q_i, K_j)) = \phi(Q_i)\varphi(K_j).
\end{aligned}
\end{equation}

To satisfy these three above requirements, here we propose the Equation~\ref{eq:sim_and_distance}, as a map from distance to similarity.

\begin{equation}
\label{eq:sim_and_distance}
\begin{aligned}
Sim(Q_i, K_j) &= \sum_{p=1}^{n}cos(\frac{\pi}{4}|\hat{Q}_{ip} - \hat{K}_{jp}|) \\
\hat{Q}_{ip} &= Tanh({Q}_{ip}) \\
\hat{K}_{ip} &= Tanh({K}_{jp}) \\
Tanh(x) &= \frac{e^x - e^{-x}}{e^x + e^{-x}} \\
\end{aligned}
\end{equation}

For any $x \in \mathbb{R}$, $-1 < Tanh(x) < 1$, and the following relation always hold.

\begin{equation}
\label{eq:sim_and_distance_2}
\begin{aligned}
& -2 < \hat{Q}_{ip} - \hat{K}_{jp} < 2 \\
\implies & 0 < \frac{\pi}{4}|\hat{Q}_{ip} - \hat{K}_{jp}| < \frac{\pi}{2}
\end{aligned}
\end{equation}

Note that $Tanh(x)$ is a monotonically increasing function, and in domain $x \in (0, \frac{\pi}{2})$, $cos(x)$ is a monotonically decreasing function, so the map function $f(x) = cos(Tanh(x))$, from distance to similarity, is monotonically decreasing. The listed first requirement is satisfied.

As $\frac{\pi}{4}|\hat{Q}_{ip} - \hat{K}_{jp}|\in(0, \frac{\pi}{2})$, $Sim(Q_i, K_j) = \sum_{p=1}^{n}cos(\frac{\pi}{4}|\hat{Q}_{ip} - \hat{K}_{jp}|) > 0$, the listed second requirement is satisfied.

Because $cos(x)$ is an even function, the absolute value symbol in Equation~\ref{eq:sim_and_distance} and~\ref{eq:sim_and_distance_2} can be removed. As a result, the similarity function can be decomposed. Let $Q_{ip}^{cos} = cos(\frac{\pi}{4}(\hat{Q}_{ip}))$, $K_{jp}^{cos} = cos(\frac{\pi}{4}(\hat{K}_{jp}))$, $Q_{ip}^{sin} = sin(\frac{\pi}{4}(\hat{Q}_{ip}))$, $K_{jp}^{sin} = sin(\frac{\pi}{4}(\hat{K}_{jp}))$, the decomposed similarity function is proved in Equation~\ref{eq:sim_and_distance_3}. The introduced third requirement, decomposed possibility, is now satisfied.
\begin{equation}
\label{eq:sim_and_distance_3}
\begin{aligned}
Sim(Q_i, K_j) &= \sum_{p=1}^{n}cos(\frac{\pi}{4}(\hat{Q}_{ip} - \hat{K}_{jp})) \\
&= \sum_{p=1}^{n}cos(\frac{\pi}{4}(\hat{Q}_{ip}))cos(\frac{\pi}{4}(\hat{K}_{jp}))\\
&+ \sum_{p=1}^{n}sin(\frac{\pi}{4}(\hat{Q}_{ip}))sin(\frac{\pi}{4}(\hat{K}_{jp})) \\
&= \sum_{p=1}^{n}Q_{ip}^{cos}K_{jp}^{cos} + \sum_{p=1}^{n}Q_{ip}^{sin}K_{jp}^{sin}
\end{aligned}
\end{equation}

Now, the $Q(Disco(K, C))$ in architecture Figure~\ref{fig:architecture} is defined as in the Equation~\ref{eq:proposed_linear attention}.

\begin{equation}
\label{eq:proposed_linear attention}
\begin{aligned}
O_i &= \sum_{j=1}^{N} {\frac{Sim(Q_i, K_j)}{\sum_{j=1}^{N} Sim(Q_i, K_j)} V_j} \\
    =& \sum_{j=1}^{N} {\frac{\sum_{p=1}^{n}cos(\frac{\pi}{4}(\hat{Q}_{ip} - \hat{K}_{jp})))}{\sum_{j=1}^{N} \sum_{p=1}^{n}cos(\frac{\pi}{4}(\hat{Q}_{ip} - \hat{K}_{jp}))} V_j} \\
    =& \sum_{j=1}^{N} {\frac{\sum_{p=1}^{n}Q_{ip}^{cos}K_{jp}^{cos} + \sum_{p=1}^{n}Q_{ip}^{sin}K_{jp}^{sin}}{\sum_{j=1}^{N}{\sum_{p=1}^{n}Q_{ip}^{cos}K_{jp}^{cos}} + \sum_{j=1}^{N}{\sum_{p=1}^{n}Q_{ip}^{sin}K_{jp}^{sin}}}V_j} \\
    =& \frac{\sum_{p=1}^{n}Q_{ip}^{cos}(\sum_{j=1}^{N}{K_{jp}^{cos}V_j}) + \sum_{p=1}^{n}Q_{ip}^{sin}(\sum_{j=1}^{N}{K_{jp}^{sin}V_j})}{\sum_{j=1}^{N}{\sum_{p=1}^{n}Q_{ip}^{cos}K_{jp}^{cos}} + \sum_{j=1}^{N}{\sum_{p=1}^{n}Q_{ip}^{sin}K_{jp}^{sin}}} \\
    &\overset{\mathrm{def}}{=} Q_{i}(Disco(K, C))
\end{aligned}
\end{equation}

Equation~\ref{eq:proposed_linear attention} is the proposed attention function, and it is an extension of the general Equation~\ref{eq:2_1} and~\ref{eq:general_sim_and_distance} with only linear complexity $O(N)$.

In our architecture in Figure~\ref{fig:architecture}, the SFC grouping Transformer is used to build group internal receptive field. Although the shifted group method is used, the receptive field is only limited expanded. With the proposed DISCO module, the voxel has a real global receptive field with only linear complexity $O(N)$.

\subsection{Channel attention module and decoder}
Let $x_{SFC} \in \mathbb{R} ^ {N \times C}$ denotes the output of SFC-Grouping Transformer module, and $x_{DISCO} \in \mathbb{R} ^ {N \times C}$ denotes the output of DISCO module. Here $N$ is the number of voxels, $C$ is the number of channels.

The $x_{SFC}$ and $x_{DISCO}$ are at first concatenated in channel dimension. The concatenated output is denoted as $x \in \mathbb{R} ^ {N \times 2C}$.

Like the similar idea in~\cite{chaneel_attention}, a  channel descriptor $w \in \mathbb{R} ^ {2C}$ is calculated by squeezing the global spatial information. 
 Let $x_{ij}$ denote the $i$-th voxel and its $j$-th channel, and $a \in \mathbb{R} ^ {2C}$ denotes the maxpooling output from $x$. For any $j$, $a_j = max(\{x_{ij} | i \in \{1, 2, 3 ... N\} \})$. The descriptor $w \in \mathbb{R} ^ {2C}$ is then calculated as the softmax output of $a$.

The channel attention module output is denoted as $o \in \mathbb{R} ^ {N \times 2C}$. For any $o_{ij}$, it is the product of the input $x_{ij}$ and the descriptor $w_j$ like

\begin{equation}
\label{eq:channel_attention}
\begin{aligned}
o_{ij} = x_{ij}w_j.
\end{aligned}
\end{equation}

The channel attention module output is then decoded by a simple multi-layer perceptron network, and the label of each point is further predicted.

\begin{table*}[h]
  \centering
  \scriptsize
  \begin{tabular}{p{3.2cm} C{0.5cm} C{0.4cm} C{0.22cm} C{0.22cm} C{0.22cm} C{0.22cm} C{0.22cm} C{0.22cm} C{0.22cm} C{0.22cm} C{0.22cm} C{0.22cm} C{0.22cm} C{0.22cm} C{0.22cm} C{0.22cm} C{0.22cm} C{0.22cm} C{0.22cm} C{0.22cm} C{0.22cm} C{0.22cm}}
    \toprule
    Method & FPS & \rotatebox{90}{mIoU($\%$)} & \rotatebox{90}{road} & \rotatebox{90}{sidewalk} & \rotatebox{90}{parking} & \rotatebox{90}{other-ground} & \rotatebox{90}{building} & \rotatebox{90}{car}  & \rotatebox{90}{truck} & \rotatebox{90}{bicycle} & \rotatebox{90}{motorcycle} &\rotatebox{90}{other-vehicle} & \rotatebox{90}{vegetation} & \rotatebox{90}{trunk} & \rotatebox{90}{terrain} & \rotatebox{90}{person} & \rotatebox{90}{bicyclist} & \rotatebox{90}{motorcyclist} & \rotatebox{90}{fence} & \rotatebox{90}{pole} & \rotatebox{90}{traffic-sign}\\
    \midrule
    PointNet~\cite{qi2017pointnet} & 10.12 & 14.6 &  61.6 & 35.7 &  15.8 & 1.4 & 41.4 & 46.3 & 0.1 & 1.3 & 0.3 & 0.8 & 31.0 & 4.6 & 17.6 & 0.2 & 0.2 & 0.0 & 12.9 & 2.4 & 3.7 \\
    PointNet++~\cite{qi2017pointnet++} & 0.06 & 20.1 &  72.0 & 41.8 &   18.7 & 5.6 & 62.3 & 53.7 & 0.9 &  1.9 &  0.2 &  0.2 & 46.5 &  13.8 &  30.0 & 0.9 &  1.0 & 0.0 & 16.9 & 6.0 & 8.9 \\
    RandLA~\cite{randLA} & 1.74 & 53.9 &  90.7 & 73.7 &   60.3 & 20.4 &  86.9 & 94.2 & 40.1 &  26.0 &   25.8 &  38.9 &  81.4 &  61.3 &  66.8 & 49.2 &  48.2 & 7.2 & 56.3 & 49.2 & 47.7 \\
    MVP-Net~\cite{Luo2022A03_mvpNEt}& \textbf{19.20} & 53.9 &  91.4 & \textbf{75.9} & 61.4 & 25.6 &  85.8 & 92.7 & 20.2 &  37.2 & 17.7 & 13.8 &  83.2 &   64.5 & 69.3 & 50.0 &  55.8 & 12.9 & 55.2 & 51.8 & 59.2 \\
    KPConv~\cite{thomas2019kpconv} & 0.88 & 58.8 &  88.8 & 72.7 &    61.3  & 31.6 & 90.5 & 96.0 & 33.4 &  30.2 & 42.5 &  44.3 &  84.8 &  69.2 &  69.1 & 61.5 &  61.6 & 11.8 & 64.2 & 56.4 & 47.4 \\
    \midrule
    RangeNet53++~\cite{milioto2019rangenet++} & 16.12 & 52.2 & 91.8 & 75.2 & 65.0 & 27.8 & 87.4 &  91.4 & 25.7 & 25.7 & 34.4 &  23.0 & 80.5 & 55.1 &  64.6 & 38.3 & 38.8 &  4.8 &  58.6 & 47.9 & 55.9 \\
    SqueezeSegV3~\cite{xu2020squeezesegv3} & 6.49 & 55.9 &  91.7 & 74.8 & 63.4 & 26.4 &  89.0 & 92.5 & 29.6 &  38.7 & 36.5 & 33.0 &  82.0 &   59.4 &   65.4 & 45.6 &  46.2 & 20.1 & 58.7 & 49.6 &  58.9 \\
    PolarNet~\cite{zhang2020polarnet} & 1.96 & 54.3 &  90.8 & 74.4 &   61.7 & 21.7 & 90.0 & 93.8 & 22.9 &  40.3 &    30.1 &  28.5 &  84.0 &  61.3 &   65.5 & 43.2 &  40.2 & 5.6 & 61.3 & 51.8 & 57.5 \\
    PMF~\cite{zhuang2021pmf} & - & 63.9 & \textbf{96.4} & 80.5 & 43.5 & 0.1 & 88.7 & 95.4 & \textbf{68.4} & 47.8 & 62.9 & \textbf{75.2} & \textbf{88.6} & \textbf{72.7} & \textbf{75.3} & \textbf{78.9} & 71.6 & 0.0 & 60.1 & \textbf{65.6} & 43.0\\
    \midrule
    SparseConv(Baseline)~\cite{3DSemanticSegmentationWithSubmanifoldSparseConvNet, yan2021JS3C-Net} & - & 61.8 & 89.9 & 72.1 & 56.5 & 29.6 & 90.5 & 94.5 & 43.5 & 51.0 & 42.4 & 31.3 & 83.9 & 67.4 & 68.3 & 60.4 & 61.3 & 41.1 & 65.6 & 57.9 & 67.7 \\
    JS3C-Net~\cite{yan2021JS3C-Net} & - & 66.0 & 88.9 & 72.1 & 61.9 & 31.9 & \textbf{92.5} & 95.8 & 54.3 & 59.3 & 52.9 & 46.0 & 84.5 & 69.8 & 67.9 & 69.5 & 65.4 & 39.9 & \textbf{70.8} & 60.7 & 38.7 \\ 
     SPVNAS~\cite{spvnas} & 0.88 & 66.4 & - & - & - & - & - & - & - & - & - & - & - & - & - & - & - & - & - & - & - \\
     Cylinder3D~\cite{Zhu_2021_cylinder3d} & 2.55 & 67.8 & 91.4 & 75.5 & \textbf{65.1} & 32.3 & 91.0 & \textbf{97.1} & 59.0 & \textbf{67.6} & 64.0 & 58.6 & 85.4 & 71.8 & 68.5 & 73.9 & 67.9 & 36.0 & 66.5 & 62.6 & 65.6\\
     
     \midrule
     LEST(ours) & 6.58 & \textbf{69.7} & 91.0 & 75.0 & 62.0 & \textbf{32.4} & 92.1 & 96.5 & 44.5 & 65.4 & \textbf{65.2} & 55.4 & 86.4 & 72.6 & 70.9 & 77.6 & \textbf{77.7} & \textbf{71.0} & 69.2 & 62.2 & \textbf{68.3} \\
 \bottomrule
 \\
  \end{tabular}
  \caption{The comparison results of our network and the other state-of-the-art point-based, projection-based and voxel-based methods on the SemanticKITTI test dataset. Note that like other methods~\cite{Zhu_2021_cylinder3d}, test time augmentation, including point cloud flip and rotation, is applied. FPS, frames per second, represents the inference speed. All these methods' speed, FPS, is measured on a TESLA V100 with GPU memory 32G. mIoU, mean intersection-over-union, is the accuracy metric over all classes.}
  \label{tab:semantickitti_result}
\end{table*}

\begin{table*}[h]
  \centering
  \scriptsize
  \begin{tabular}{p{2.2cm} C{0.4cm} C{0.22cm} C{0.22cm} C{0.22cm} C{0.22cm} C{0.22cm} C{0.22cm} C{0.22cm} C{0.22cm} C{0.22cm} C{0.22cm} C{0.22cm} C{0.22cm} C{0.22cm} C{0.22cm} C{0.22cm} C{0.22cm}}
    \toprule
    Method & \rotatebox{90}{mIoU($\%$)} & \rotatebox{90}{barrier} & \rotatebox{90}{bicycle} & \rotatebox{90}{bus} & \rotatebox{90}{car} & \rotatebox{90}{construction} & \rotatebox{90}{motorcycle}  & \rotatebox{90}{pedestrian} & \rotatebox{90}{traffic-cone} & \rotatebox{90}{trailer} &\rotatebox{90}{truck} & \rotatebox{90}{driveable} & \rotatebox{90}{other} & \rotatebox{90}{sidewalk} & \rotatebox{90}{terrain} & \rotatebox{90}{manmade} & \rotatebox{90}{vegetation}\\
    \midrule
    RangeNet53++~\cite{milioto2019rangenet++} & 65.5 & 66.0 & 21.3 & 77.2 & 80.9 & 30.2 & 66.8 & 69.6 & 52.1 & 54.2 & 72.3 & 94.1 & 66.6 & 63.5 & 70.1 & 83.1 & 79.8 \\
    PolarNet~\cite{zhang2020polarnet} & 71.0 & 74.7 & 28.2 & 85.3 & 90.9 & 35.1 & 77.5 & 71.3 & 58.8 & 57.4 & 76.1 & 96.5 & 71.1 & 74.7 & 74.0 & 87.3 & 85.7 \\
    Salsanext~\cite{cortinhal2020salsanext} & 72.2 & 74.8 & 34.1 & 85.9 & 88.4 & 42.2 & 72.4 & 72.2 & 63.1 & 61.3 & 76.5 & 96.0 & 70.8 & 71.2 & 71.5 & 86.7 & 84.4 \\
    Cylinder3D~\cite{Zhu_2021_cylinder3d} & 76.1 & 76.4 & 40.3 & 91.2 & \textbf{93.8} & 51.3 & 78.0 & 78.9 & 64.9 & 62.1 & \textbf{84.4} & 96.8 & \textbf{71.6} & 76.4 & 75.4 & \textbf{90.5} & 87.4 \\
    PMF~\cite{zhuang2021pmf} & 76.9 & 74.1 & 46.6 & 89.8 & 92.1 & \textbf{57.0} & 77.7 & 80.9 & \textbf{70.9} & \textbf{64.6} & 82.9 & 95.5 & 73.3 & 73.6 & 74.8 & 89.4 & 87.7\\
    \midrule 
    LEST(ours) & \textbf{77.1} & \textbf{79.1} & \textbf{47.4} & \textbf{91.8} & 87.5 & 49.2 & \textbf{86.1} & \textbf{82.4} & 70.5 & 58.6 & 80.6 & \textbf{96.9} & 71.4 & \textbf{76.8} & \textbf{77.1} & 90.2 & \textbf{88.5} \\
 \bottomrule
 \\
  \end{tabular}
  \caption{The comparison results of our network and the other state-of-the-art methods on the nuScenes validation dataset.}
  \label{tab:nuScenes_result}
\end{table*}

\section{Experiments}

In this section, we provide the experiments results at first. The model is trained and evaluated on the two large-scale LiDAR-based semantic segmentation datasets, SemanticKITTI~\cite{behley2019semantickitti} and nuScenes~\cite{nuscenes_od, fong2022nuscenes_panoptic}. The results are then compared with other state-of-the-art approaches and the performance differences are analyzed. Finally, a series of ablation studies are conducted to validate the proposed modules.

\subsection{Datasets and evaluation metric}
\subsubsection{SemanticKITTI}
SemanticKITTI is a large-scale LiDAR-based semantic segmentation dataset. The point cloud data is derived from the KITTI~\cite{geiger2012kitti} Vision Odometry Benchmark. Point-wise annotations are labeled for the complete 360\textdegree \ field-of-view of the employed Velodyne-HDLE64 LiDAR. This dataset consisits of 22 sequences. The 00-07, 09, and 10 sequences are commonly used for training, and the 08 sequence is used for validation. The rest 11-21 sequences are used as test set. After officially merging similar classes and ignoring classes with too few points, 19 classes are evaluated in the single scan perception task.

\subsubsection{nuScenes}
The nuScenes dataset is a multimodal dataset for autonomous driving. It comprises 1000 scenes of 20 seconds duration data from a 32-beams LiDAR sensor. This dataset is officially split into a training set and a validation set. In our work, the model is trained on the training set and evaluated on the validation set. Similarly to SemanticKITTI, classes with too few points are ignored and similar classes are merged during training and evaluation. In total, 16 classes are trained and evaluated in our approach.

\subsubsection{Evaluation metric}
The mean intersection-over-union (mIoU) over all classes is widely used as evaluation metric. It is formulated as 
\begin{equation}
\label{eq:miou}
\begin{aligned}
mIoU = \frac{1}{C}\sum_{i=1}^{C}\frac{TP_i}{TP_i + FP_i + FN_i}.
\end{aligned}
\end{equation}

In Equation~\ref{eq:miou}, $C$ is the number of the classes. $TP_i$, $FP_i$, $FN_i$ are the true positive, false positive, false negative predictions for class $i$. 
\subsection{Results on SemanticKITTI}
In this section, our approach is compared with the other LiDAR-only state-of-the-art approaches, including point-based method, projection-based method and voxel-based method. The results on the SemanticKITTI test set is shown in Table~\ref{tab:semantickitti_result}.

Compared to all the other point-based~\cite{qi2017pointnet, qi2017pointnet++, randLA, Luo2022A03_mvpNEt, thomas2019kpconv}, projection-based~\cite{milioto2019rangenet++, xu2020squeezesegv3, zhang2020polarnet} and voxel-based method~\cite{3DSemanticSegmentationWithSubmanifoldSparseConvNet, yan2021JS3C-Net, yan2021JS3C-Net, spvnas, Zhu_2021_cylinder3d}, our approach has significant performance improvement in terms of mIoU.

Note that the current voxel-based methods are actually sparse-3D-convolution-plus method. JS3C-Net~\cite{yan2021JS3C-Net} uses sparse 3D convolution and takes advantage of multiple-frames information. SPVNAS~\cite{spvnas} uses neural architecture search (NAS) method to find out the best sparse 3D convolution network architecture. Cylinder3D~\cite{Zhu_2021_cylinder3d} uses the cylindrical partitions instead of the normal 3D voxels, and process these cylindrical partitions with the vanilla sparse 3D convolution.

However, our method can be used as an alternative to the sparse 3D convolution. Compared to the SparseConv baseline~\cite{3DSemanticSegmentationWithSubmanifoldSparseConvNet, yan2021JS3C-Net}, our method has a 7.9\% absolute mIoU improvement. 

The other methods such as using multiple-frames information, NAS method, cylindrical partition and image information~\cite{zhuang2021pmf}, can be further applied to our current method in future work to improve the performance. Our method uses only the normal 3D voxels and single frame, which is LiDAR-independent and more compatible to the other state-of-the-art methods~\cite{yin2021center_point} in multi-tasks learning like 3D object detection.

\subsection{Results on nuScenes}
In this section, our method is compared with the other methods on nuScenes validation set. The result is shown in Table~\ref{tab:nuScenes_result}. Our proposed LEST model performs better than the other methods, especially on the small object like bicycle, motorcycle and pedestrian. Compared to the SemanticKITTI equipped with 64-beams LiDAR, nuScenes with 32-beams LiDAR has fewer points per scan. As a result, our model LEST can have larger perceptive field in SFC grouping branch with the same limited GPU resource.

\begin{figure*}[h]
\begin{center}
\includegraphics[width=1.0\linewidth]{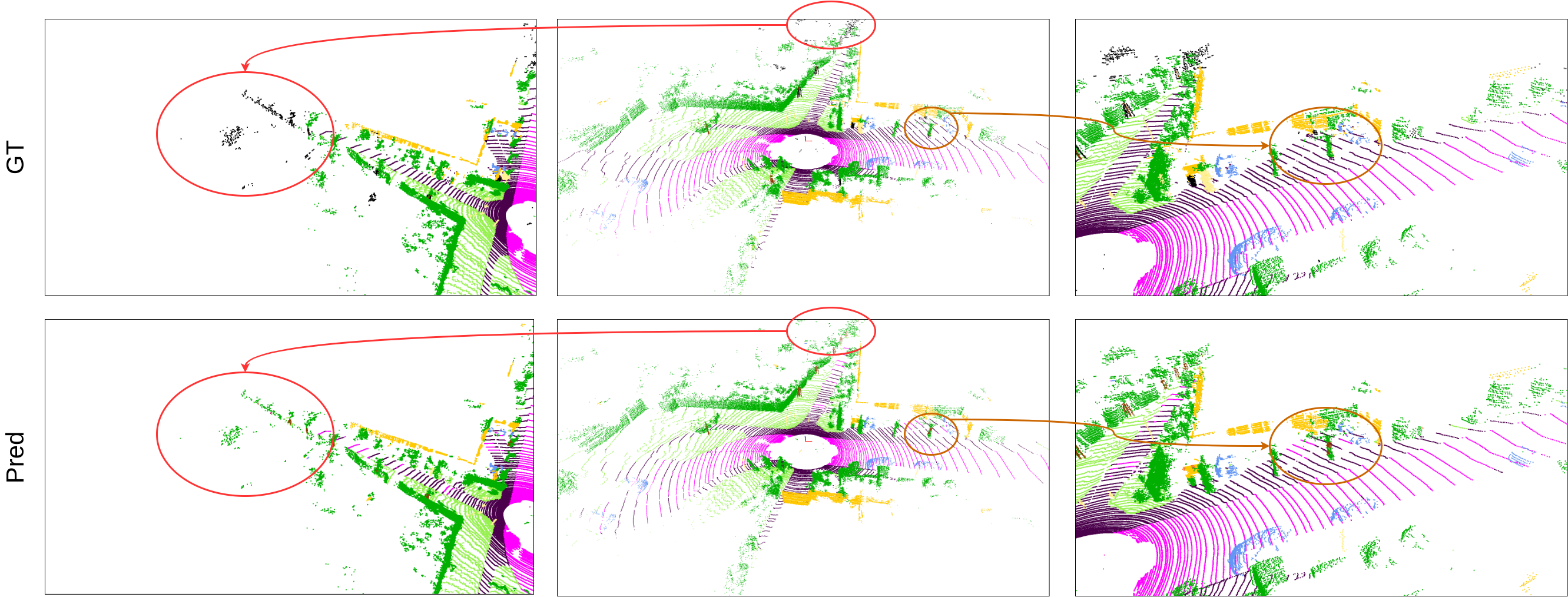}
\end{center}
\caption{A qualitative comparison of our method prediction with the Ground-truth. The first row GT is the Ground-truth. The second row Pred is our model LEST prediction. In the 1st column. the GT has no labeling in the far range, but our method predicts the points as trees and terrains. Unlike the dense 2D method~\cite{lang2019pointpillars}, ~\cite{zhang2020polarnet}, our method can have unlimited prediction range and predict the whole points. The 3d column shows a failure case of our method. The sidewalk points are wrongly classified as the road, and the trees are wrongly classified as the trunk. The reason is that, though we use DISCO module to reduce the geometry information compact caused by SFC grouping, there is still some minor errors from SFC grouping.}
\label{fig:qualitative_results}
\end{figure*}

\subsection{Qualitative results}
Figure~\ref{fig:qualitative_results} shows the qualitative results of our model's prediction and the ground-truth. Unlike some works~\cite{zhang2020polarnet, Zhu_2021_cylinder3d}, our method is fully sparse and can have unlimited range in training and prediction. As a result, even the unlabeled points can be classified. The qualitative results also shows one limitation of our method: the local points cannot be correctly classified as the same sometimes. The reason is the geometry information loss caused by SFC even we use shifted SFC and DISCO.

\subsection{Ablation studies}
In this section, all the proposed components will be validated. The training dataset is the nuScenes training set and the validation dataset is the nuScenes validation set.

Our proposed LEST model consists of the shifted SFC-Grouping Transformer, the DISCO module and a channel attention module. The shifted SFC-Grouping is validated by comparing with removing the shifted SFG-Grouping module completely or using only single SFG-Grouping. The DISCO module is validated by removing it or replacing it with the other state-of-the-art linear Transformers like CosFormer~\cite{zhen2022cosformer} and kernel function based Linear Transformer~\cite{katharopoulos2020linear_transformer}. The channel attention module is validated by removing it and directly concatenating the features from multiple branches. The results are listed in Table~\ref{tab:ablation}.

From the ablation experiments 1st row in Table~\ref{tab:ablation}, it can be observed that removing the SFC-Grouping branch and using only the DISCO module has poor performance. One reason is the Low-Rank Bottleneck~\cite{bhojanapalli2020low_rank_bottleneck, dong2021attention_is_not_all_you_need, katharopoulos2020linear_transformer} in Transformer. In the vanilla Transformer, let $ x \in \mathbb{R} ^ {N \times C}$ denotes $N$ tokens with features dimension $C$, and the learn-able matrices are $W_Q \in \mathbb{R} ^ {C \times D}$, $W_K \in \mathbb{R} ^ {C \times D}$. In ~\cite{bhojanapalli2020low_rank_bottleneck} it is proved that, for any $x$ and one  arbitrary positive column stochastic matrix $P\in \mathbb{R} ^ {N \times N}$, if $D \geqslant N$, there always exist the matrix $W_Q$, $W_K$ satisfying that $softmax(\frac{(xW_Q)(xW_K)^T}{\sqrt{D}}) = P$. If $D < N$, there exist $X$ and $P$ such that this equation does not hold for all $W_Q$ and $W_K$. In linear Transformer scenario, $D \ll N$, and the Low-Rank problem is worse.

The 2nd row in Table~\ref{tab:ablation} shows that the used shifted grouping method performs better than using only one group. The reason is that the receptive field is expanded in shifted method. The 3rd-5th rows show that our proposed linear Transformer, DISCO, performs better than the other state-of-the-art linear Transformer~\cite{zhen2022cosformer}~\cite{katharopoulos2020linear_transformer}. The 6th row shows that the attention module is necessary to aggregate the features from multiple branches.

\begin{table}[h]
	\centering
        \scriptsize
	\begin{tabular}{l c  c c}
        SFC-Grouping & Linear Transformer & Channel Attention & mIoU \\
        \midrule
        \xmark & DISCO & \checkmark & 56.7 \\
	single SFC-Grouping & DISCO & \checkmark & 73.5 \\
	\midrule
        shifted SFC-Grouping & \xmark & \checkmark & 74.8 \\
        shifted SFC-Grouping & CosFormer~\cite{zhen2022cosformer} & \checkmark & 75.4 \\
        shifted SFC-Grouping & Kernal Function~\cite{katharopoulos2020linear_transformer} & \checkmark & 76.1 \\
	\midrule
        shifted SFC-Grouping & DISCO & \xmark & 76.2 \\
        \midrule
	shifted SFC-Grouping & DISCO & \checkmark & 77.1(original) \\
	\bottomrule \\
	\end{tabular}
\caption{Ablation experiments based on the original framework}
\label{tab:ablation}
\end{table}

\section{Conclusion and outlook}
In this paper, we propose a novel pure Transformer architecture, LEST, in LiDAR-based semantic segmentation tasks. LEST consists of the SFC-Grouping module and the DISCO module, a distance-based Transformer with linear complexity. Compared to the other semantic segmentation models, LEST performs impressively and can be regarded as an alternative to the widely used sparse 3D convolution. 

With the proposed pure Transformer architecture, we would like to reduce the domain gap between 3D Computer Vision and the Natural Language Processing (NLP) field. The proposed linear Transformer, DISCO, can be also used and evaluated in NLP field in future work, especially in the long range sequences tasks.

\end{document}